\documentclass[runningheads,a4paper]{llncs}
\usepackage{geometry}
\usepackage{graphicx}
\usepackage{epstopdf}
\usepackage{amsmath}
\usepackage{subfigure}
\usepackage{float}
\usepackage{epstopdf}
\usepackage{color}
\usepackage{epsfig}
\usepackage{caption}
\usepackage{color}
\begin{document}
\title{Object Tracking in Hyperspectral Videos with Convolutional Features and Kernelized Correlation Filter}
%
%
\author{Kun Qian\inst{1} \and Jun Zhou\inst{2} \and Fengchao Xiong\inst{3} \and Huixin Zhou\inst{1} \and Juan Du\inst{1}}

\authorrunning{Kun Qian et al.}
%
\institute{Lab of Optoelectronic Imaging and Image Processing, Xidian University\\ Xi'an, P.R. China\\
\and School of Information and Communication Technology, Griffith University\\ Brisbane, Australia\\
\and College of Computer Science, Zhejiang University\\ Hangzhou, P.R. China
}
\maketitle              
\begin{abstract}
Target tracking in hyperspectral videos is a new research topic. In this paper, a novel method based on convolutional network and Kernelized Correlation Filter (KCF) framework is presented for tracking objects of interest in hyperspectral videos. We extract a set of normalized three-dimensional cubes from the target region as fixed convolution filters which contain spectral information surrounding a target. The feature maps generated by convolutional operations are combined to form a three-dimensional representation of an object, thereby providing effective encoding of local spectral-spatial information. We show that a simple two-layer convolutional networks is sufficient to learn robust representations without the need of offline training with a large dataset. In the tracking step, KCF is adopted to distinguish targets from neighboring environment. Experimental results demonstrate that the proposed method performs well on sample hyperspectral videos, and outperforms several state-of-the-art methods tested on grayscale and color videos in the same scene.

\end{abstract}
\section{Introduction}

Hyperspectral imaging plays an important role in remote sensing as it provides hundreds of contiguous, narrow spectral bands~\cite{HSI}. With the advantage of rich spectral information, hyperspectral images (HSIs) have been widely used in many applications involving image classification~\cite{a1} and segmentation~\cite{a2}, such as land cover detection and mining. However, to the best of our knowledge, there is very little work focusing on hyperspectral video processing. The main reason is that it is difficult to capture hyperspectral videos with low speed imaging devices. It is not until the last a couple of years that low cost hyperspectral video cameras become available, making it possible to collect hyperspectral videos at a high frame rate.

In this paper, we introduce one of the very first work on object tracking in hyperspectral videos. Object tracking is an important research topic in computer vision and multimedia. Most tracking methods~\cite{CN,ESK,KCF,STC,DLT} were developed on grayscale or RGB videos. Discriminative correlation filter (DCF)~\cite{CN,ESK,KCF,MOSSE} based framework explores supervised visual object tracking. The DCF trains the filters very efficiently in the frequency domain via fast Discrete Fourier transform (DFT). It learns a correlation filter to localize the object in consecutive frames. The learned filter is applied to estimate the target location by calculating the maximum response. Bolme et al. introduced the minimum output sum of squared error filter (MOSSE) tracker~\cite{MOSSE} which utilizes grayscale features and achieves an impressive speed in tracking application. Other features used for tracking include the incorporation of kernels and histogram of gradient (HOG) features~\cite{KCF}, the addition of color name features~\cite{CN}, adaptive scale~\cite{a4}, and the integration of deep learning features~\cite{a5}. The kernelized correlation filter (KCF) method~\cite{KCF} circularly shifts the training samples and exploits the advantage of multichannel HOG features with the kernel trick. Zhang et al. proposed the Spatio-Temporal Context (STC)~\cite{STC} tracker, which explores the correlation filter in terms of the probability theory, and utilized the dense sampling to track the object of interest.

In recent years, deep learning methods have shown success in object tracking~\cite{a6,a7,a8}. Several works~\cite{a9,a10} have combined deep learning with the correlation filter based framework. Instead of using hand-crafted features such as HOG, the DCF trackers use features automatically learnt by convolutional neural network (CNN). This significantly improves the robustness of the tracking. Zhang et al. proposed the lightweight convolutional network based tracker (CNT)~\cite{CNT} which has a simple architecture and yet effectively constructs a robust representation. This tracker demonstrates that a two-layer CNN without pooling and training process can obtain competitive results on a benchmark dataset with 50 challenging videos, and outperforms the first deep learning based tracker (DLT)~\cite{DLT} by a large margin.

In this paper, we propose a novel convolutional feature based tracker for hyperspectral video processing. The videos were captured by a hyperspectral camera of 14 bands in the range of 470-620nm. We first defined convolution filters from a set of normalized three-dimensional cubes surrounding a target. The convolutional operations generate a set of feature maps that are combined to form a three-dimensional representation of an object, which is used in the tracking process. In the tracking step, KCF is adopted to distinguish targets from neighboring environment. We extend the KCF method so it can cope with hyperspectral data.

The remainder of this paper is organized as follows. In Section~\ref{sec:method}, we first present the convolutional feature for hyperspectral images. Then, we briefly describe the KCF tracker, and how it can be extended for multichannel convolutional features for hyperspectral tracking. In Section~\ref{sec:experiments}, experimental results are presented to verify the performance of the proposed method on hyperspectral video sequences. Our method is also compared with the state-of-art methods on grayscale and RGB videos of the same scene. Finally, conclusions are drawn in Section~\ref{sec:conclusions}.

\section{The Proposed Tracking Algorithm}\label{sec:method}
In this section, we describe the details of the proposed method. We first introduce convolutional features in the 3D spectral-spatial domain. Then we describe the KCF method and its extension to hyperspectral data.

\subsection{Convolutional Features for Hyperspectral Target}
\begin{figure}[htbp]
\centering
  \includegraphics[width=\textwidth]{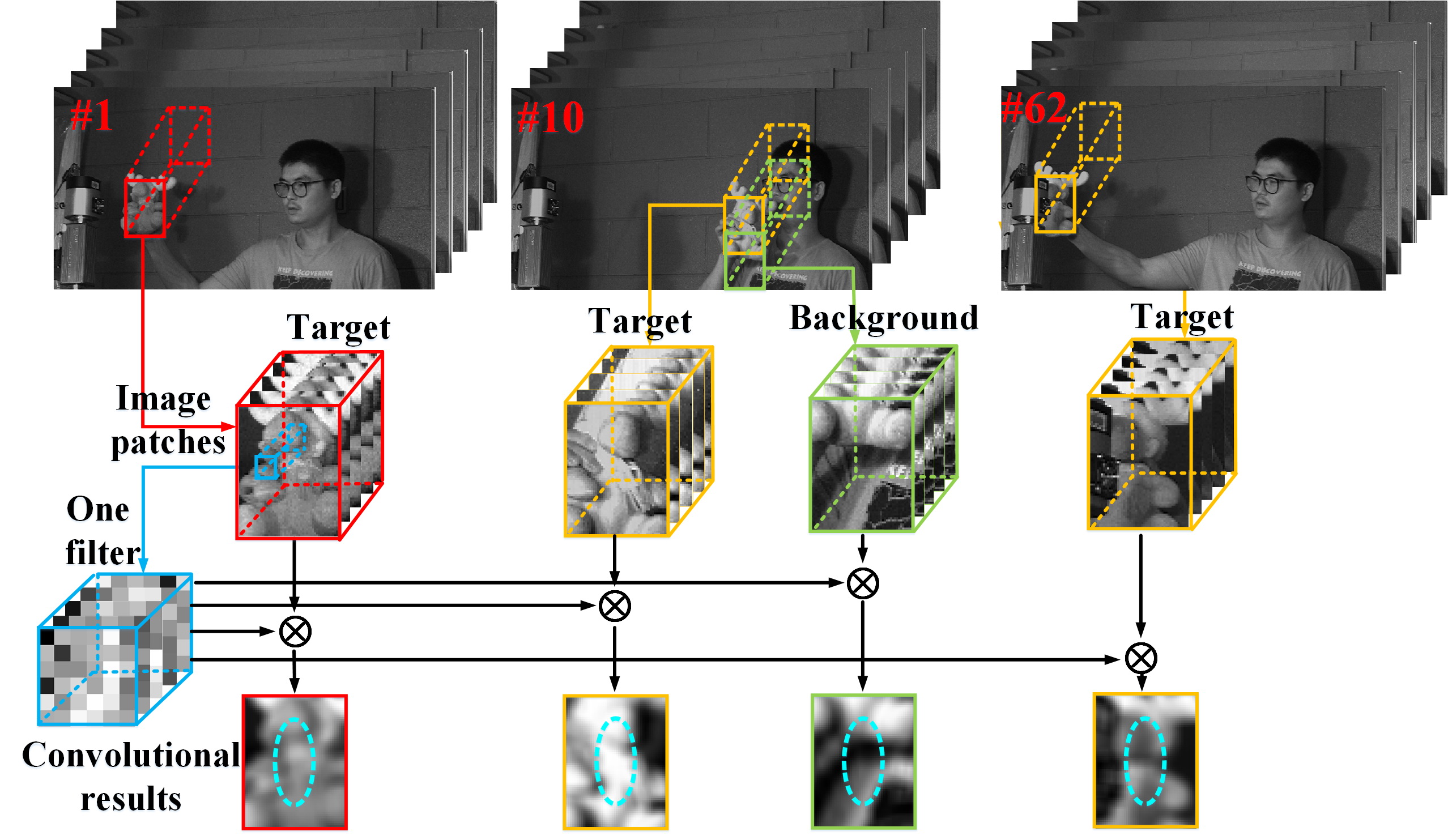}
    \caption{3D convolution process. }\label{conv}
\end{figure}

\begin{figure}[htbp]
\centering
  \includegraphics[width=\textwidth]{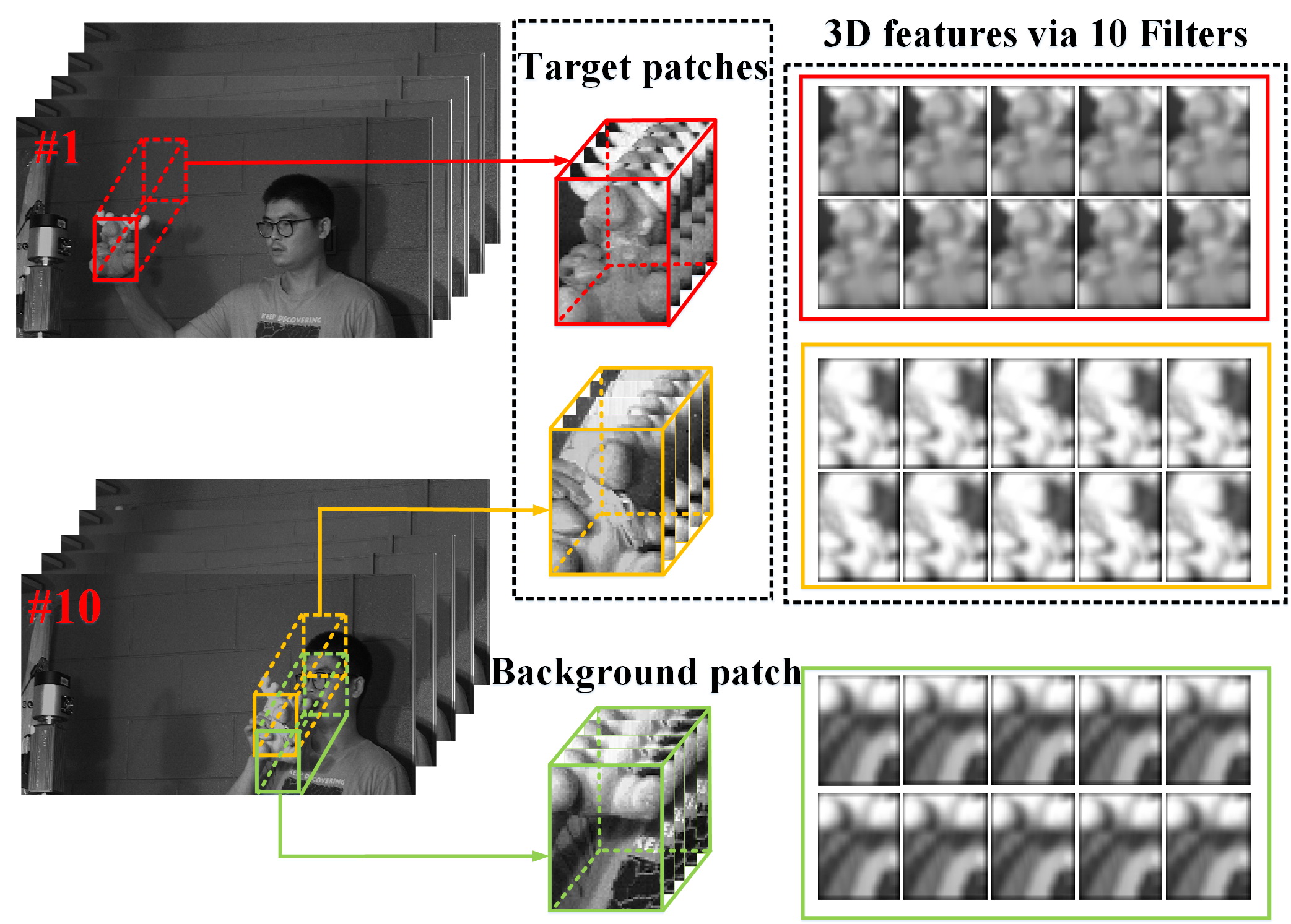}
    \caption{Stacked features. }\label{stacking}
\end{figure}

Motivated by the success of convolution network on visual tracking \cite{CNT}, we utilize this method to extract the local hyperspectral information. Given a target template, the proposed hierarchical representation architecture contains two steps. First, local features which contain spectral information are extracted from a bank of three-dimensional filters convolving with the input image at each position. Then, these features are stacked together to form a three-dimensional representation. This feature extraction process is shown in Fig.~\ref{conv} and Fig.~\ref{stacking}.

The image patchs in Fig.~\ref{conv} are generated from local hyperspectral image cube $I \in R^{n \times n \times d}$, where $n$ and $d$ denote patch size and the number of spectral bands, respectively. A set of overlapping local image patchs $Y_i (i=1,2,...,l)\in R^{w \times w \times d}$ centered at each pixel position is densely sampled inside the image patch $I$ through a sliding window of size $w \times w$, where $l=n \times n$. In the first frame of the video, several filters $f_j(j=1,2,...,d)(d<l)$ are selected randomly from $Y_i (i=1,2,...,l)$, the responses on the image patch $I$ are denoted with feature maps $S_j \in R^{n \times n}$, which can be expressed as
\begin{equation}\label{1}
S_j=I \bigotimes f_j
\end{equation}
where $\bigotimes$ is the convolution operator.

Fig.~\ref{conv} shows that the 3D filter, which is localized, can extract local structural features for the hyperspectral cubes. Furthermore, convolutional results of three target templates (at the bottom of Fig. \ref{conv}) are similar in geometric layout, which demonstrates that the local filter $f_j$ is effective in extracting the target features despite their appearance variation. For negative templates (see the third image patch in the bottom row of Fig.~\ref{conv}), its convolutional result are very different from the target templates. As shown in Fig.~\ref{stacking}, 3D features generated by 10 filters have similar properties. Therefore, the convolution results and the generated features represent the inner structure of the tracking target.

\subsection{Tracking Framework}

The output of the filtering operation are stacked to form a three-dimensional representation. This can be considered as the multichannel feature which is required as the input to the kernelized correlation filter (KCF) tracker. In this section, we firstly briefly describe the KCF tracker~\cite{KCF}, and then extend it to use the effective features introduced in the above section.

\subsubsection{The KCF Tracker}

Our approach is built on the KCF tracker which has achieved impressive results on Visual Tracker Benchmarks~\cite{VOT}. The key of the KCF algorithm is to train a classifier through a ridge regression model, whose objective function is represented as
\begin{equation}\label{2}
\min_{w} ( (w^Tx-y)^2 +\lambda \| w \|^2)
\end{equation}
where  $y$, $\lambda$ and $w$ represent the regression value, regularization parameter and regression coefficient, respectively.

The KCF approach densely samples a circulant sample matrix $X=C(x)$, where $C(x)$ denotes the circulant operation based on the first row $x$ (i.e., base sample). This matrix can be decomposed into
\begin{equation}\label{3}
C(x)=F \cdot diag (Fx) \cdot F^H
\end{equation}
where $F$ and $diag(\cdot)$ denote the DFT matrix and the diagonalization function, respectively. $F^H$ is the Hermitian transpose of $F$, which is a constant.

Improved via the kernel trick, coefficient $w$ is mapped to a high-dimensional feature space, i.e., $w=\alpha^T \varphi (x)$, where $\varphi(\cdot)$ means the mapping function and $\alpha$ is a new coefficient. Then, the coefficient $\alpha$ can be formulated as
\begin{equation}\label{4}
\alpha =(K+\lambda I)^{-1}y
\end{equation}
\begin{equation}\label{5}
F\alpha =((Fk^{xx})+\lambda)^{-1}(Fy)
\end{equation}
where kernel matrix $K$ is also a circulant matrix with $k^{xx}$ denoting the first row. In the current frame, $y$ represents a prior and can be modeled as $y=b\exp(-|D/\sigma_1|^{\beta})$, where $\exp(\cdot)$ denotes the exponential function, and $b$ is a normalization constant. $D$ denotes the Euclidean distance between the target and a pixel in the neighborhood. $\sigma_1$ and $\beta$ represent a scale parameter and a shape parameter, respectively.

In Eq.~(\ref{5}), $k$ can be computed based on the Gaussian function, i.e.,
\begin{equation}\label{6}
k^{xz}=\exp(-\frac{1}{\sigma^2}(\|x\|^2+\|z\|^2)-2F^{-1}((Fx) \otimes (Fz ))).
\end{equation}

Subsequently, the object tracking task is transformed to a detection problem. The image patch $z$ of the current frame at the same target location is treated as the testing base sample, therefore, the reponse map is expressed as:
\begin{equation}\label{7}
f(z)=F^{-1}((Fk^{xz})\otimes (F{\alpha}))
\end{equation}
where $x$ and $\alpha$ are learnt before the current frame. An intuitive description is that the reponse $f(z)$ is a linear combination of the neighboring kernel value $k^{xz}$ with the weighted coefficient $\alpha$.

\subsubsection{Multichannel Convolutional Features of HSI}
Suppose the multichannel representation $x$ (which has been reshaped to one row matrix, i.e., vector) in the current frame is composed of $x=[x_1,x_2,...,x_d]$, where $x_d$ denotes the $d$-th target representation. Since the kernels are based on the dot-product, which can be computed by summing the individual dot-products for each channel, Eq.~(\ref{6}) can be rewritten with the multichannel representation $z$ in the next frame as
\begin{equation}\label{8}
k^{xz}=\exp(-\frac{1}{\sigma^2}(\|x\|^2+\|z\|^2)-2F^{-1}(\sum_{d}(F{x_d}) \otimes (Fz_d)  ))
\end{equation}
Therefore, the 3D stacked convolutional features can be seen as multichannel features referring to a pixel of the target object in KCF.

\section{Experimental Results and Analysis}~\label{sec:experiments}
In this section, we introduce the dataset used for the experiments, and provide details on the experimental setting, results, and comparison with alternatives.

\subsection{Experimental Dataset}
We performed experiments on nine image sequences. They are named as $apple-Gray$, $apple-RGB$, $apple-HSI$, $deer-Gray$, $deer-RGB$, $deer-HSI$, $people-Gray$, $People-RGB$, $People-HSI$, respectively. The sequences contain three scenes and each scene has three videos corresponding to grayscale, color, and hyperspectral format, respectively. The color scene and hyperspectral scene are the same, which were captured using a Nikon D600 camera and a Photonfocus or an Ximea hyperspectral camera. These two types of cameras were put side by side when capturing the videos. The hyperspectral cameras captured frames of 16 bands with active range of 460-630nm at 30 frames per second. After spectral calibration, the HSI is transformed into a three-dimensional data cube with 14 channels for the Photofocus camera or 11 channels for the Ximea camera. The grayscale video is formed by band image at 490nm of the HSI sequences. Therefore, the grayscale sequences are the same as the HSI sequences in the size and number of the frames, the video content, and the target location.

\begin{table}
\caption{Summary of video sequences.}\label{tab1}
\centering
\resizebox{\textwidth}{!}{
\begin{tabular}{|c|c|c|c|c|c|}
\hline
Sequence &  No. of Frames & Image size & Target size & No. of Bands & Description  \\
\hline
$apple-Gray$ & 182 & 512$\times$272 & 32$\times$30 & 1 & OCC, FM and BC \\
$apple-RGB$ & 422 & 1980$\times$1080 & 133$\times$123 & 3 & OCC and BC \\
$apple-HSI$ & 182 & 512$\times$272 & 32$\times$30 & 14 & OCC, FM and BC \\
$deer-Gray$ & 114 & 512$\times$272 & 50$\times$65 & 1 & OCC, OV and FM \\
$deer-RGB$ & 230 & 1980$\times$1080 & 170$\times$240 & 3 & OCC and OV \\
$deer-HSI$ & 114 & 512$\times$272 & 50$\times$65 & 14 & OCC, OV and FM \\
$people-Gray$ & 641 & 512$\times$272 & 45$\times$70 & 1 &  OCC, BC, IPR and DEF \\
$people-RGB$ & 676 & 1980$\times$1080 & 180$\times$280 & 3 & OCC, BC, IPR and DEF \\
$people-HSI$ & 641 & 512$\times$272 & 45$\times$70 & 14 & OCC, BC, IPR and DEF \\
\hline
\end{tabular}
}
\end{table}

\subsection{Experimental Setup}
To better analyze the strength and weakness of the tracking algorithm, we considered 6 attributes~\cite{VOT} based on different challenging factors including background clutters (BC), out of view (OV), in-plane rotation (IPR), fast motion (FM), Deformation (DEF), and Occlusion (OCC), which are summarized in Table~\ref{tab1}.

\begin{figure}[htb!]
\centering
    \subfigure[Sampled tracking results on RGB sequence.\label{a}]{
     \includegraphics[width=0.32\textwidth,trim=0 0 0 0,clip]{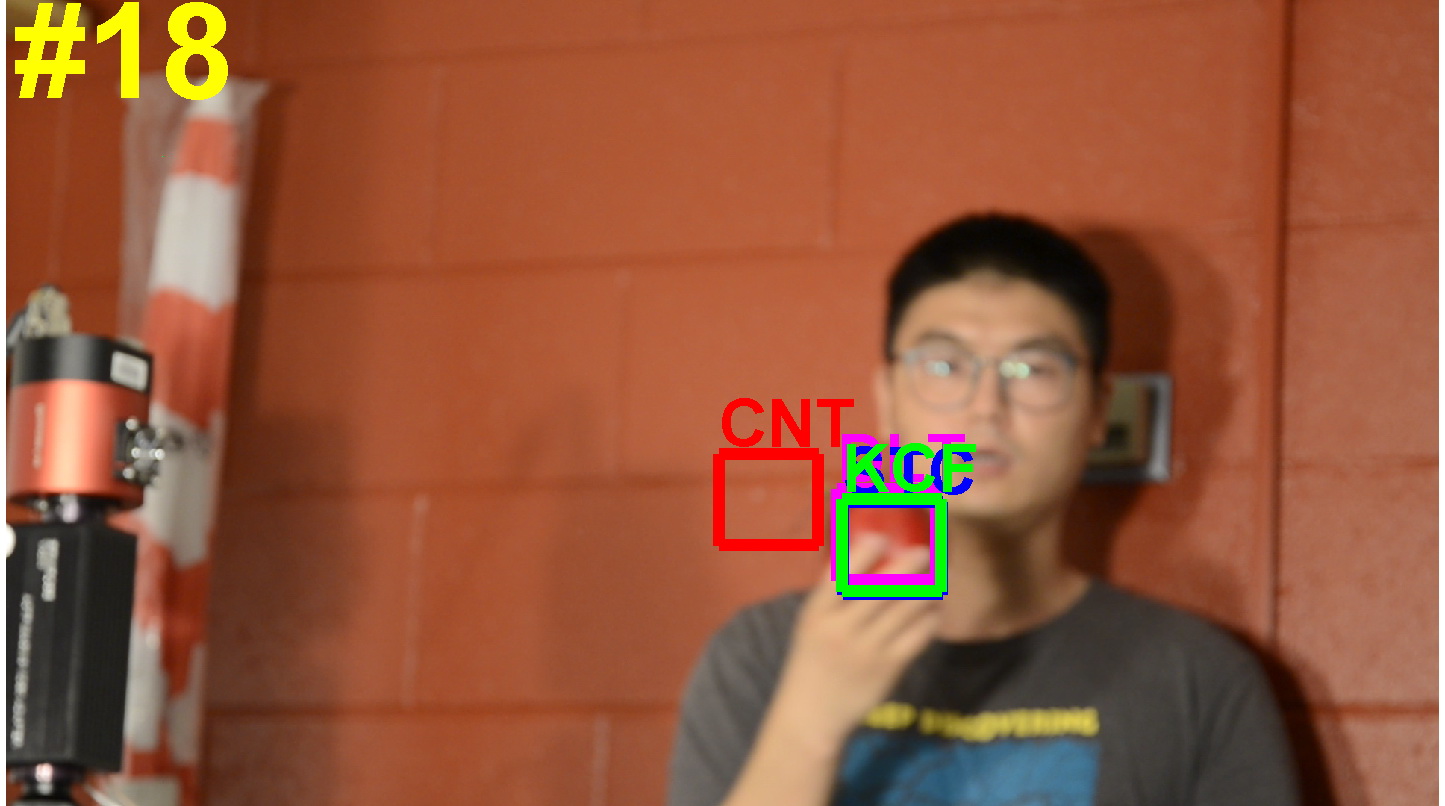}
 \includegraphics[width=0.32\textwidth,trim=0 0 0 0,clip]{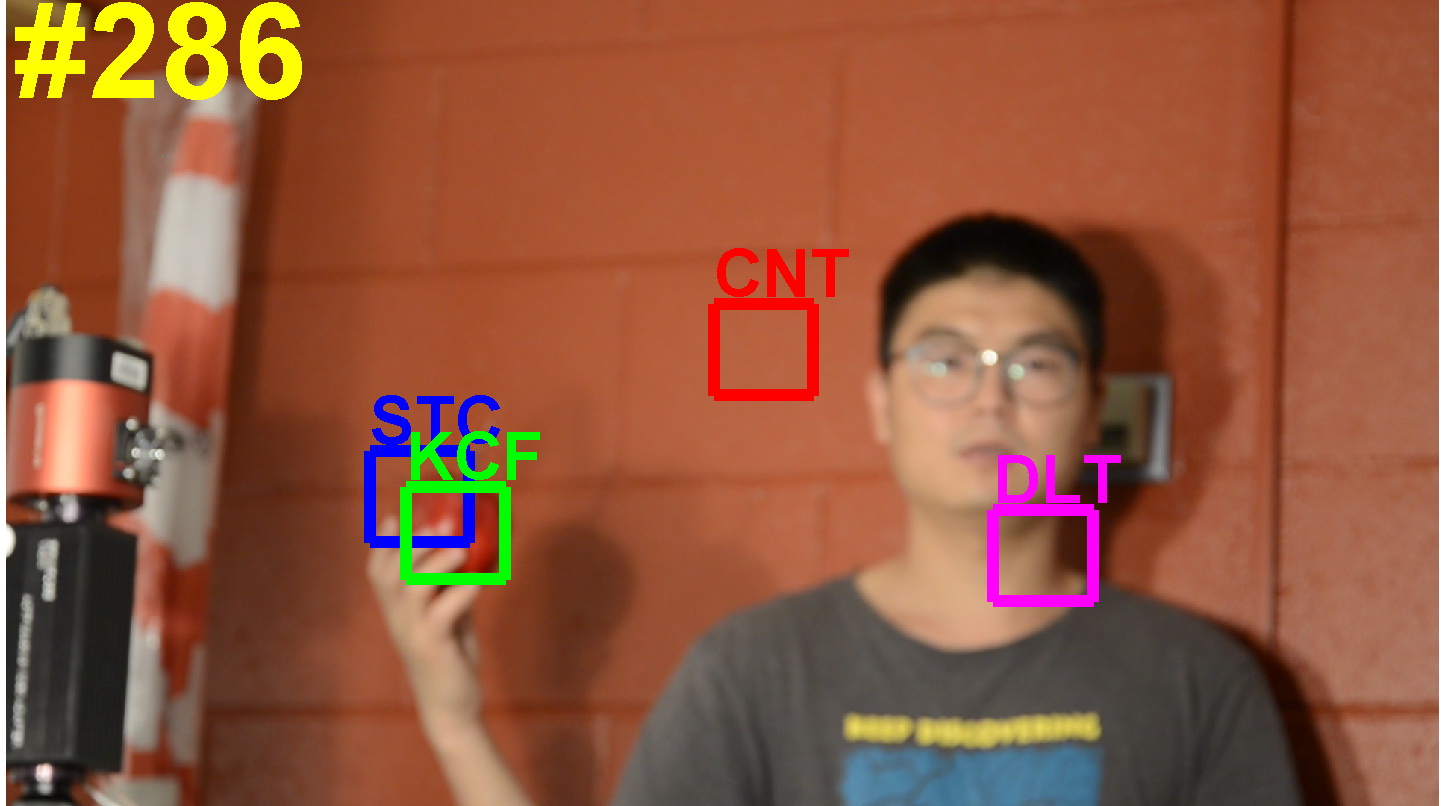}
 \includegraphics[width=0.32\textwidth,trim=0 0 0 0 10,clip]{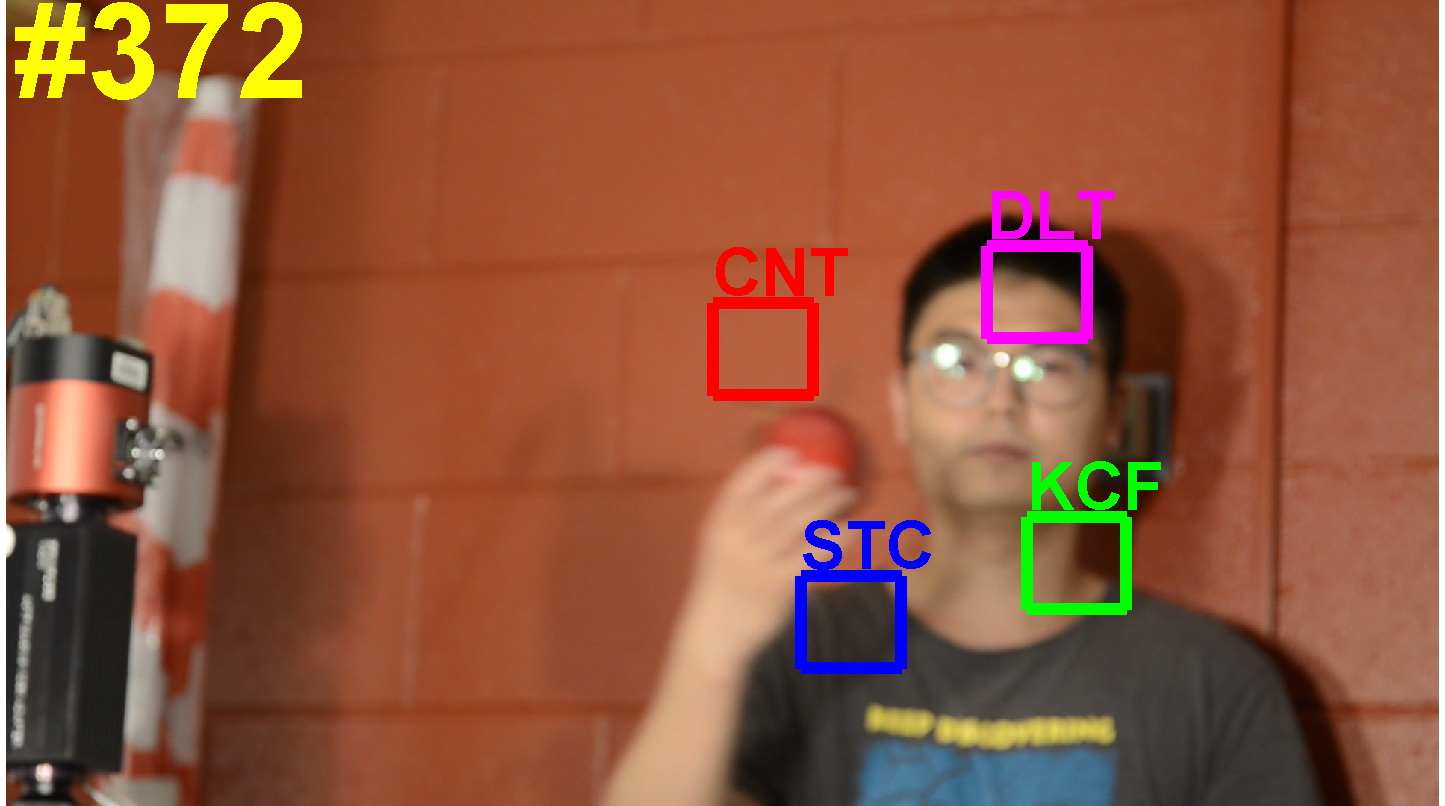}
 }
  \subfigure[Sampled tracking results on grayscale and HSI sequences.\label{b}]{
      \includegraphics[width=0.32\textwidth,trim=0 0 0 0,clip]{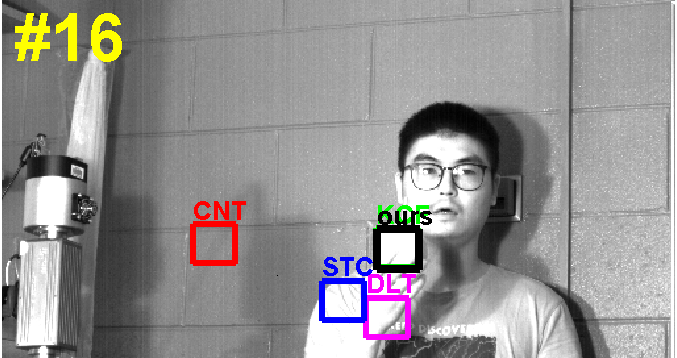}
 \includegraphics[width=0.32\textwidth,trim=0 0 0 0,clip]{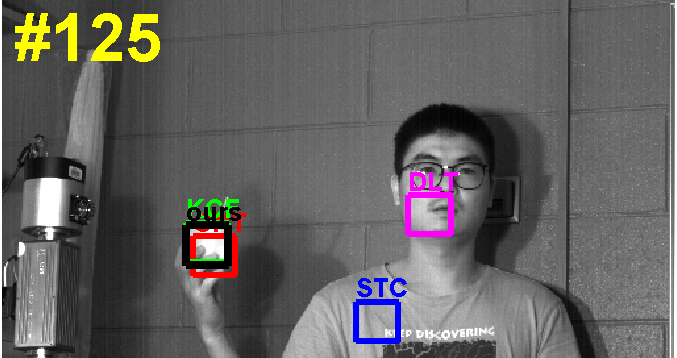}
 \includegraphics[width=0.32\textwidth,trim=0 0 0 0 10,clip]{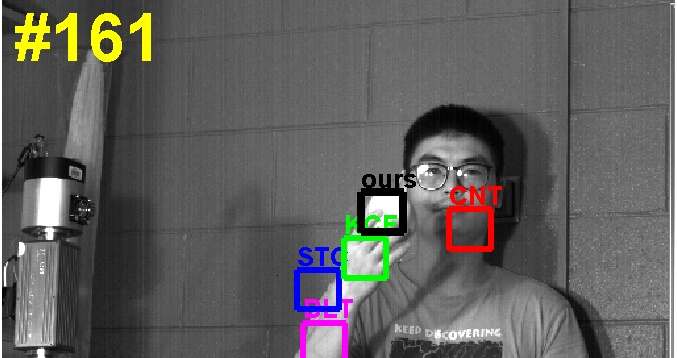}
 }
      \subfigure{
     \includegraphics[width=0.5\textwidth,trim=0 0 0 0,clip]{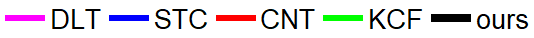}
 }
     \caption{Qualitative results on the apple sequence.}\label{apple}
\end{figure}

\begin{figure}[htb!]
\centering
        \subfigure[Sampled tracking results on RGB sequence.]{
     \includegraphics[width=0.32\textwidth,trim=0 0 0 0,clip]{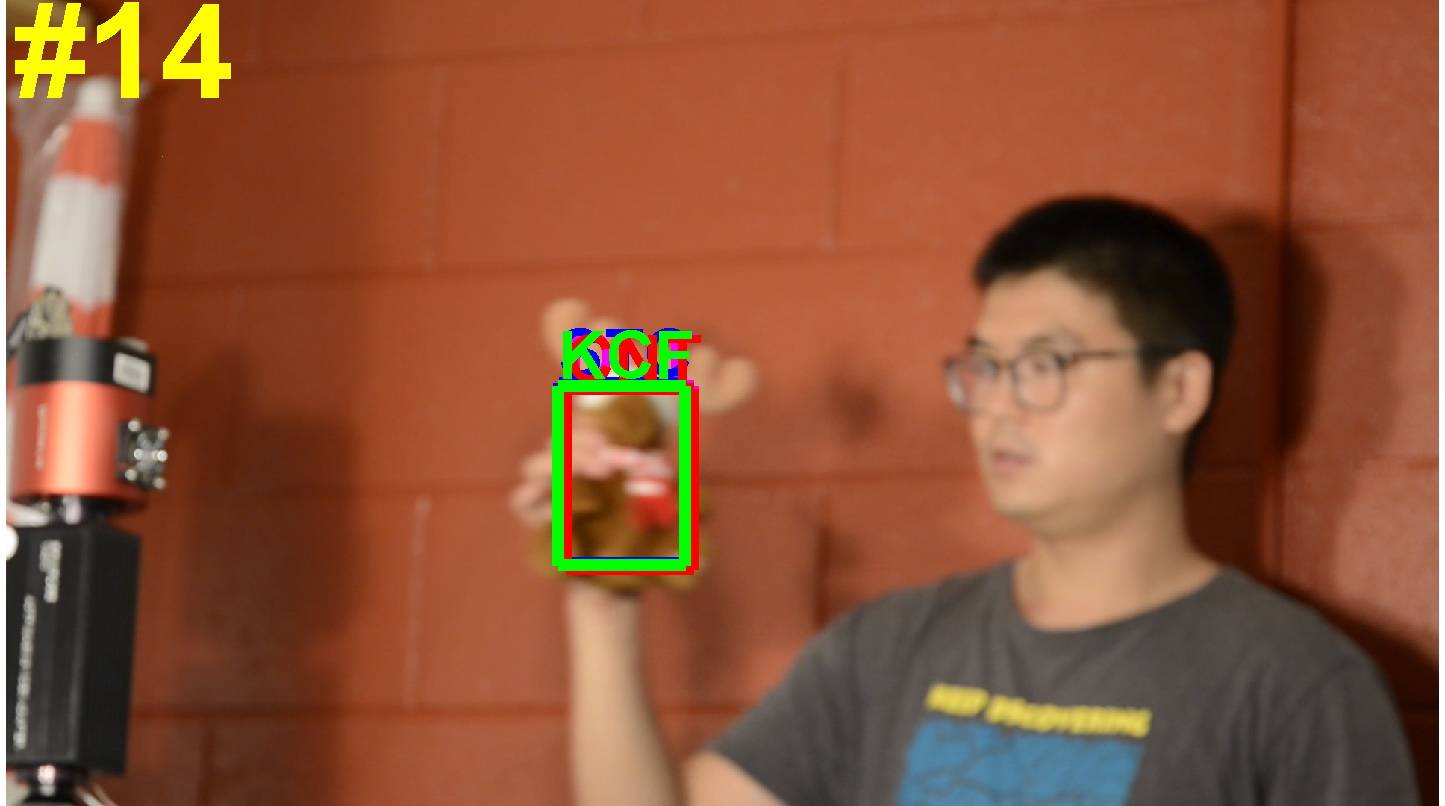}
 \includegraphics[width=0.32\textwidth,trim=0 0 0 0,clip]{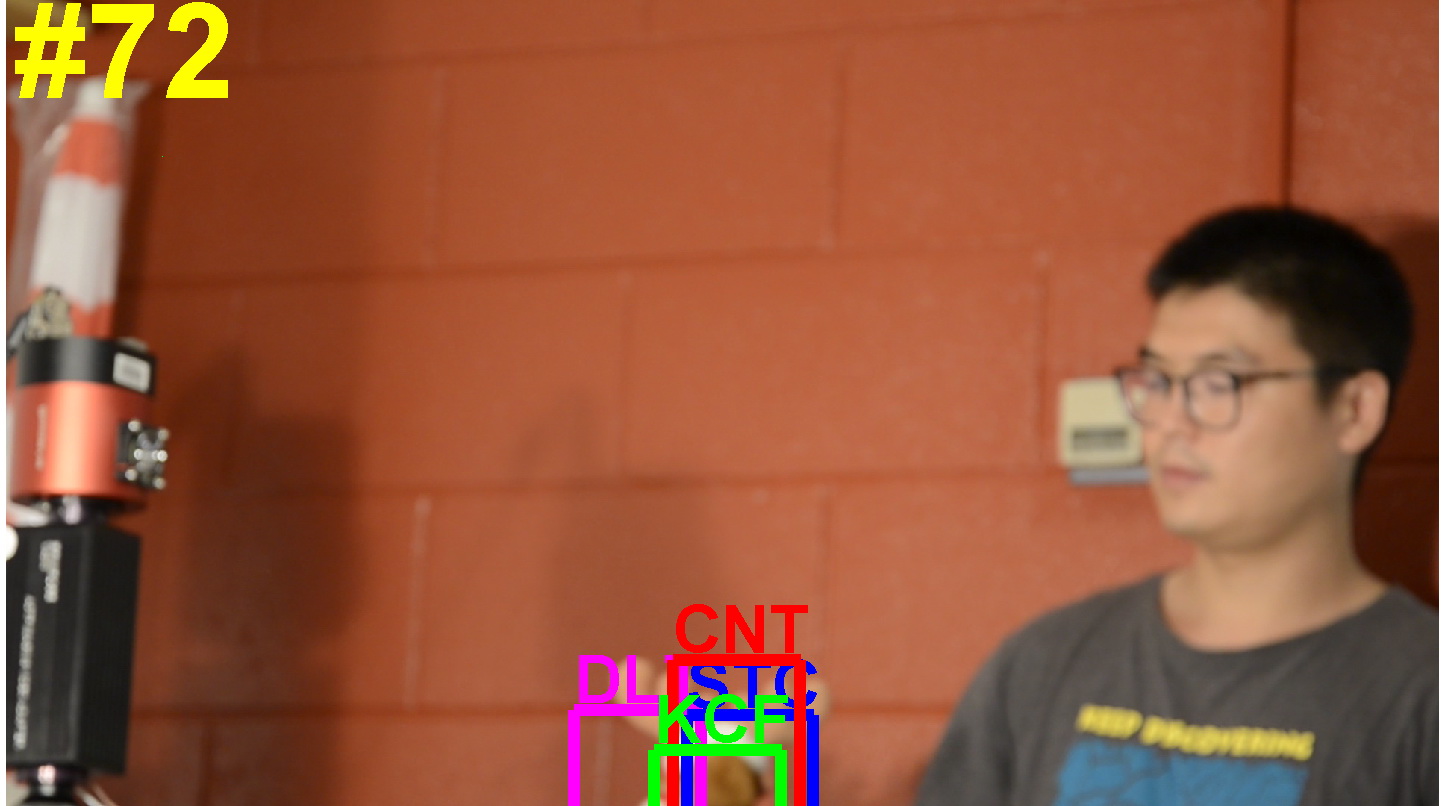}
 \includegraphics[width=0.32\textwidth,trim=0 0 0 0 10,clip]{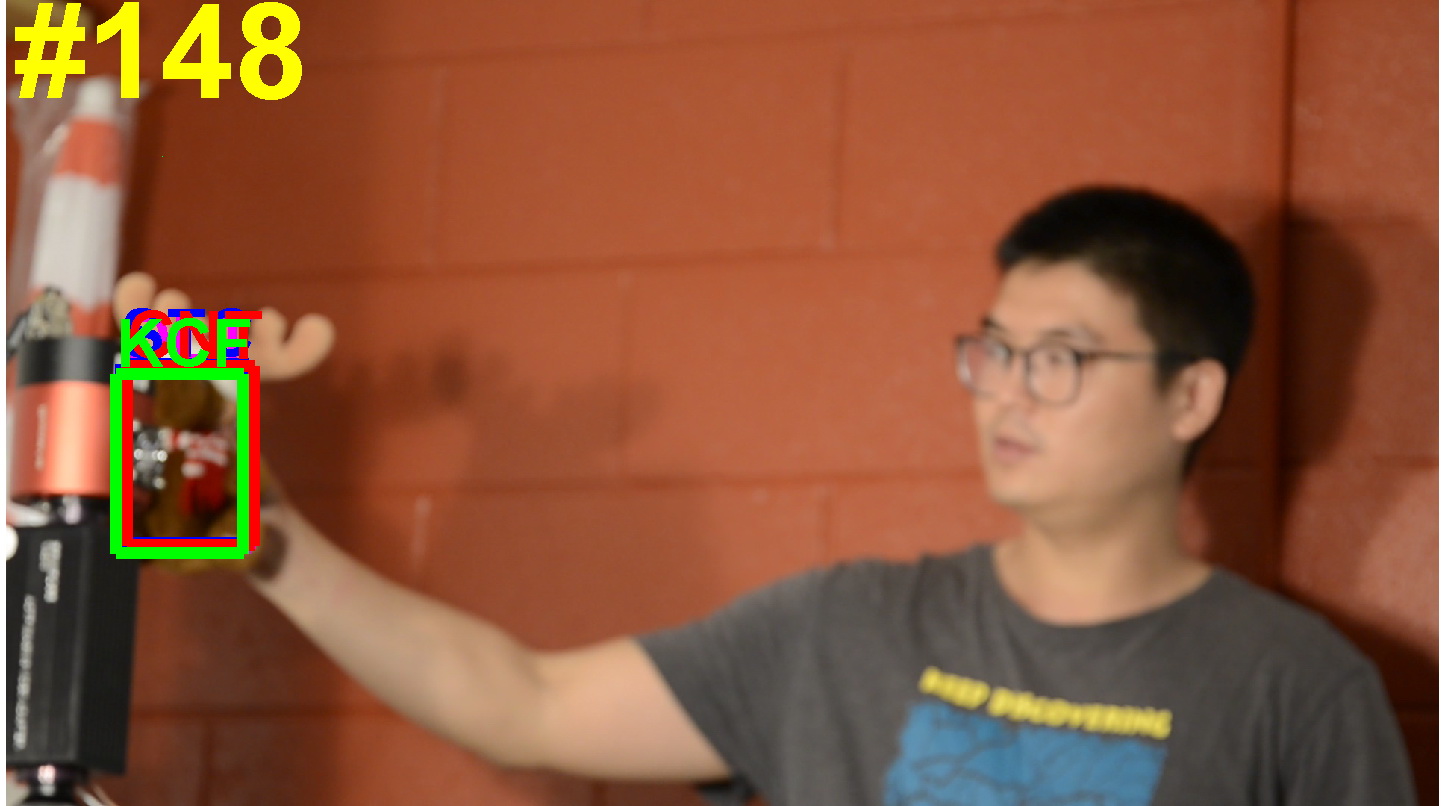}
 }
\subfigure[Sampled tracking results on grayscale and HSI sequences.]{
      \includegraphics[width=0.32\textwidth,trim=0 0 0 0,clip]{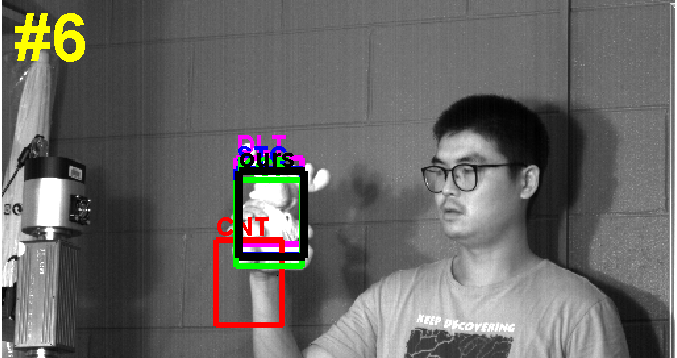}
 \includegraphics[width=0.32\textwidth,trim=0 0 0 0,clip]{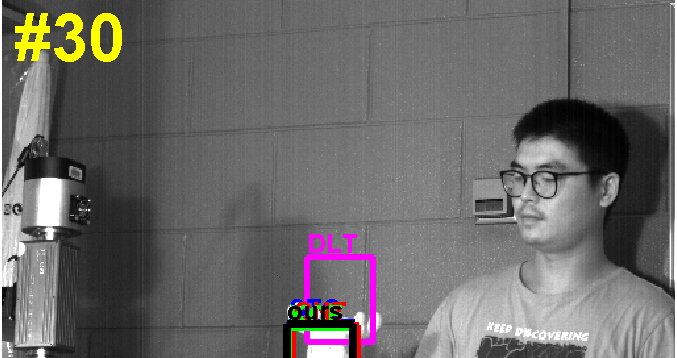}
 \includegraphics[width=0.32\textwidth,trim=0 0 0 0 10,clip]{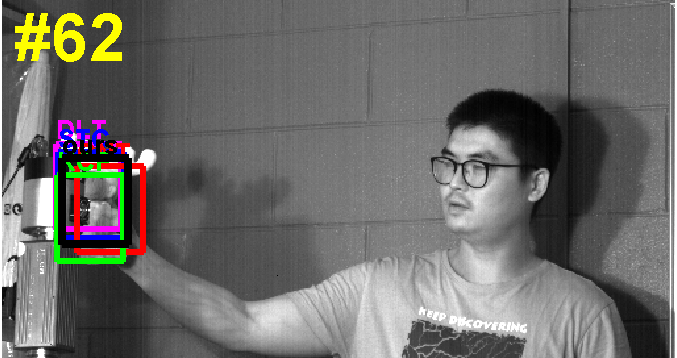}
 }
   \subfigure{
     \includegraphics[width=0.5\textwidth,trim=0 0 0 0,clip]{show//bar1.png}
 }
       \caption{Qualitative results of the deer sequence.}\label{deer}
\end{figure}

\begin{figure}[htb!]
\centering
     \subfigure[Sampled tracking results on RGB sequence.]{
    \includegraphics[width=0.32\textwidth,trim=0 0 0 0,clip]{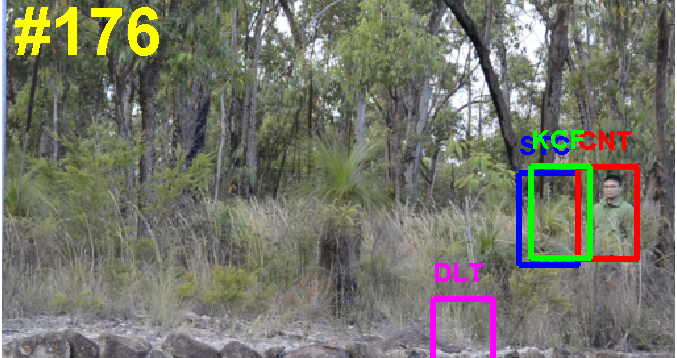}
 \includegraphics[width=0.32\textwidth,trim=0 0 0 0,clip]{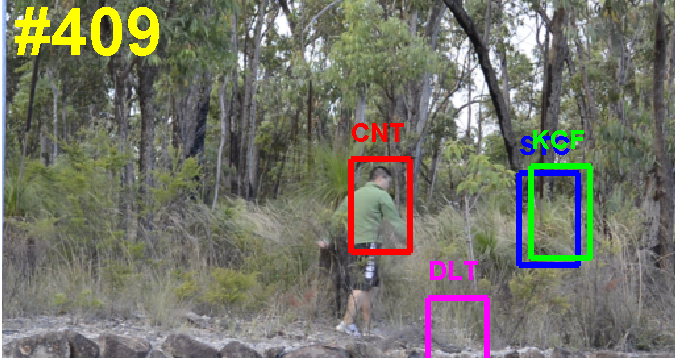}
 \includegraphics[width=0.32\textwidth,trim=0 0 0 0 10,clip]{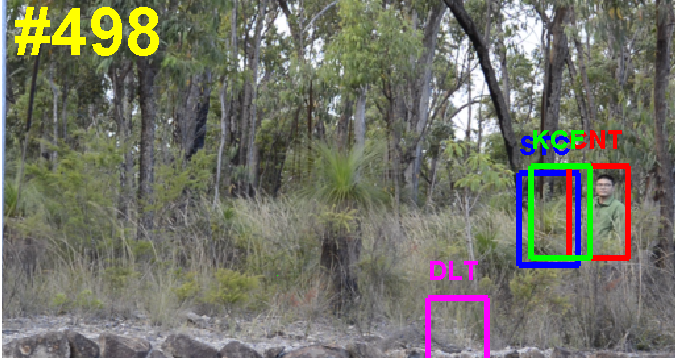}
 }
   \subfigure[Sampled tracking results on grayscale and HSI sequences.]{
       \includegraphics[width=0.32\textwidth,trim=0 0 0 0,clip]{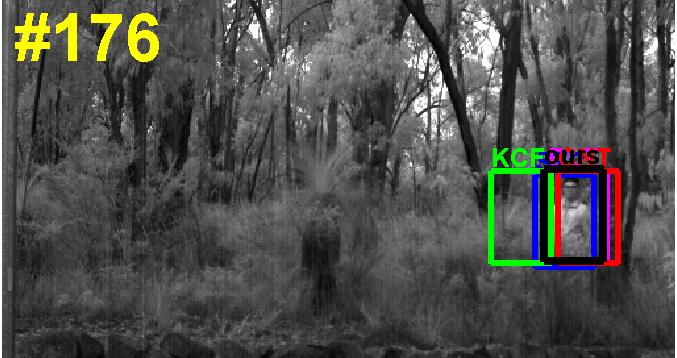}
 \includegraphics[width=0.32\textwidth,trim=0 0 0 0,clip]{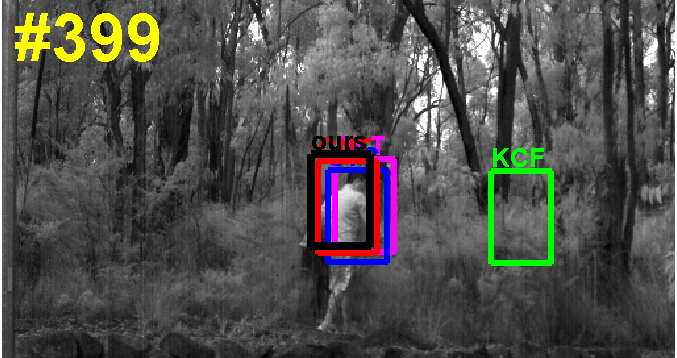}
 \includegraphics[width=0.32\textwidth,trim=0 0 0 0 10,clip]{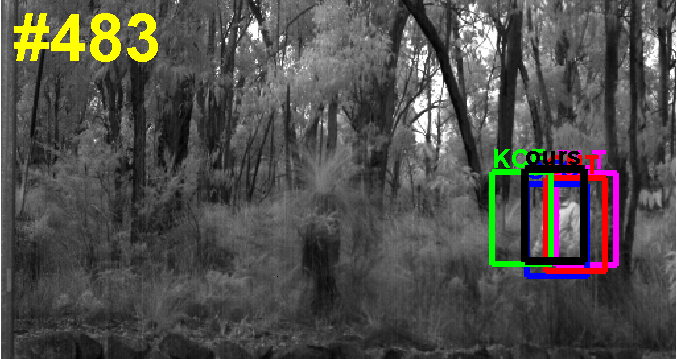}
 }
 \subfigure{
     \includegraphics[width=0.5\textwidth,trim=0 0 0 0,clip]{show//bar1.png}
 }
     \caption{Qualitative results of the people sequence.}\label{people}
\end{figure}

The proposed convolutional network based hyperspectral tracking (CNHT) method was implemented in MATLAB and ran at 1 frame per second on a PC with Intel i7-7700HQ (2.8 GHz) and 32 GB RAM. To validate the performance of the CNHT approach, we compared it with some state-of-the-art algorithms, including deep network based trackers DLT~\cite{DLT} and CNT~\cite{CNT}, and correlation filter based trackers STC~\cite{STC} and KCF~\cite{KCF}. The experimental results of the comparison are shown in Figs.~\ref{apple}-\ref{people}. For convenience, we display the results on grayscale and hyperspectral sequences in the same images as their scene are identical.

For four comparative methods, we only changed the parameter on the search scope (e.g., in STC, the search scope is fixed at 6 times of the target size), in order to adapt to fast motion of the object. In our tracker, the state of the target (i.e., size and location) in the first frame was given by the ground truth, which is carefully manually labelled. The size of the filter was set to 6$\times$6$\times$14 ($w$=6, $d$=14), the number of filters was set to a small number of 10 for high speed tracking. The size of the base sample was set as 0.2-3 times of the initial target size, in order to handle fast motion. The other parameters with respect to the KCF method remain unchanged as in the original paper.

\subsection{Qualitative Comparison}

\subsubsection{Background Clutters}
Fig. \ref{apple} and Fig. \ref{people} show some screenshots of the tracking results in sequences where the background and the target have similar color in the RGB images. In the $apple-RGB$ sequence, the color of the apple and its neighbourhood are red. The CNT method undergoes large drift in the entire sequence. The DLT, STC and KCF track the target well at the beginning of the sequence (e.g. $\#$18), but lose the target at the final stage (e.g. $\#$372). The tracking result of the KCF method on the grayscale video is more accurate, which is shown in Fig. \ref{b}. Utilizing the spectral information, our tracker is the only one that performs well on the entire sequence. The target people in the $people-RGB$ sequence is wearing a green jumper which is similar to the color of the plants. The CNT tracks the object stably, even in the $people-Gray$ sequence. The DLT tracker drifts away from the target from the beginning to the end. The STC and KCF approaches lock on parts of background when the people walks in front of the tree (e.g. $\#$176). Furthermore, the KCF faces the same problem as in the grayscale image. However, the convolutional network based KCF approach handles color similarity well thanks to the fact that it exploits the characteristic of hyperspectral features.

\subsubsection{Partial Occlusion}
The targets in all sequences contain partial occlusion. In Fig. \ref{apple}, the apple is frequently occluded by the fingers. In the gray sequences, the KCF and CNT methods are able to re-detect the object when the target re-appears in the screen (e.g., $\#125$).  In Fig. \ref{deer}, the deer is partial occluded by the camera (e.g., $\#$62 of the grayscale sequence and $\#$148 of the RGB sequence). All trackers achieve favorable results because targets of interest are large compared with the size of frames and have different appearance from the background. However, the target moves out of the screen at $\#$72 of the RGB sequence or $\#$30 of the grayscale sequence, in which frame the DLT method drifts to the background. In Fig. \ref{people}, the location estimation of the people is possibly disturbed by the thick bush (e.g., $\#$176 of either grayscale or RGB video). The KCF method does not performs well (e.g., $\#$483 of either grayscale or $\#$498 of RGB video). Nevertheless, the proposed method obtains a stable tracking target on the hyperspectral video with much better accuracy than the alternatives on the grayscale video and the RGB video.

\subsection{Quantitative Comparison}
\begin{figure}[t]
\centering
 \subfigure[Precision plot for all sequences.]{
     \includegraphics[width=0.47\textwidth]{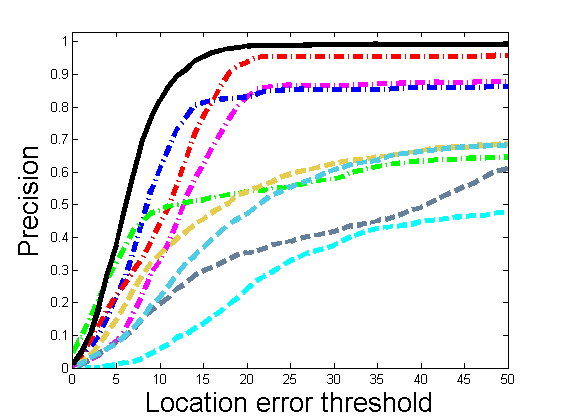}
 }
       \subfigure[Precision plot for apple sequences.]{
    \includegraphics[width=0.47\textwidth,trim=00 0 00 0,clip]{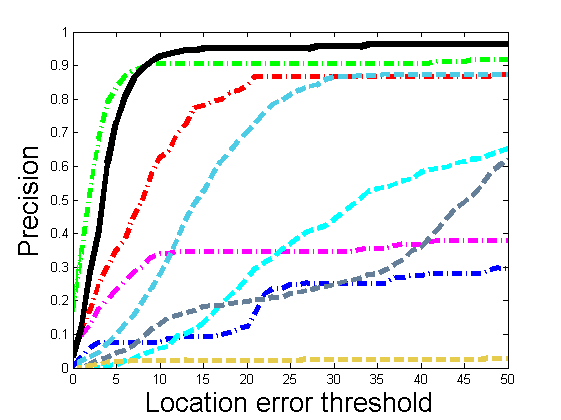}
 }
   \subfigure[Precision plot for deer sequences.]{
      \includegraphics[width=0.47\textwidth,trim=0 0 0 0,clip]{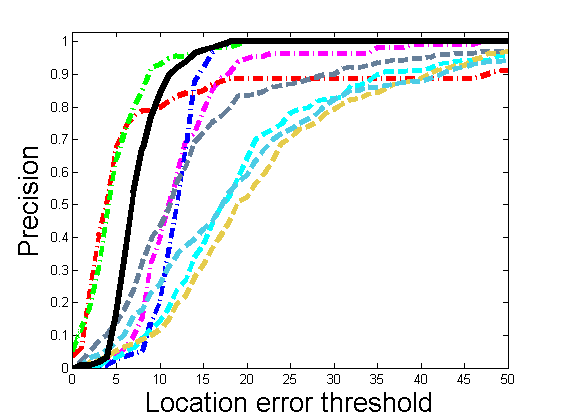}
 }
 \subfigure[Precision plot for people sequences.]{
     \includegraphics[width=0.47\textwidth]{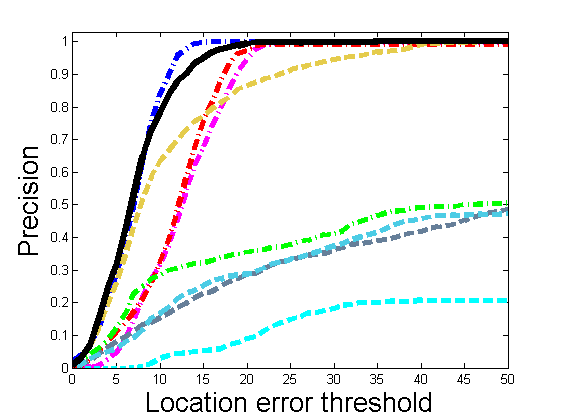}
 }
\subfigure{
     \includegraphics[width=0.9\textwidth]{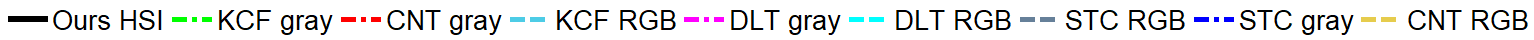}
 }
     \caption{precision plot.}\label{precision plot}
\end{figure}

\begin{table}[t]
\caption{Precision and FPS}\label{tab2}
\centering
\resizebox{0.7\textwidth}{!}{
\begin{tabular}{|c|c|c|c|}
\hline
Algorithm & Video Type & Mean Precision (20px) & Mean FPS \\
\hline
DLT & Gray & 80.4$\%$ & 0.8/20 (CPU/GPU) \\
STC & Gray & 82.8$\%$ & 365 \\
CNT & Gray & 92.7$\%$ & 0.5 \\
KCF & Gray & 53.6$\%$ & 278 \\
DLT & RGB & 21.5$\%$ &  0.6/10 (CPU/GPU) \\
STC & RGB & 34.6$\%$ & 13 \\
CNT & RGB & 53.1$\%$ & 0.5 \\
KCF & RGB & 45.8$\%$ & 65 \\
CNHT & HSI & 98.2$\%$ & 1 \\
\hline
\end{tabular}
}
\end{table}

Fig. \ref{precision plot} shows the performance of all tracking algorithms in terms of precision which is defined as the ratio of successful frames whose tracking output is within the given threshold (in pixels) from the ground-truth, measured by the center distance between bounding boxes. The precision of the proposed algorithm on the HSI sequence ranks the highest (0.982), which is followed by the CNT (0.927) on grayscale sequences. The CNHT method takes advantages of the KCF method, hyperspectral information, and convolution method. Thus, it outperforms the KCF method, which uses only one band of the hyperspectral video, by 83$\%$. The precision of STC and CNT on apple sequences are much lower than those over the deer and people sequences. This is because the apple is small and moves fast. More importantly, it has similar color as the background. As shown in Table~\ref{tab2}, the proposed CNHT method runs at 1 frame per second, which is acceptable in consideration of the multiple bands in hyperspectral videos. The algorithm efficiency can be improved in the future via running on GPUs which increase the speed of DLT method by at least 16 times.

\section{Conclusions}~\label{sec:conclusions}
In this paper, we introduce a convolutional feature based kernerlized correlation filter approach for hyperspectral video tracking. The hyperspectral features are extracted via two-layer convolutional network. They provide discriminative information and can be used as multichannel features for the KCF tracking framework. The experimental results demonstrate that the presented method performs well in a hyperspectral dataset. This lays the foundation for developing future hyperspectral tracking methods.
%
%
%

\end{document}